\documentclass[twoside,11pt]{article}

\usepackage{jmlr2e}

\usepackage{times} 			
\usepackage{graphicx} 		
\usepackage{subfigure} 
\usepackage{natbib}			
\usepackage{multicol}
\usepackage{algorithm}		
\usepackage{algorithmic}	
\usepackage[utf8]{inputenc} 
\usepackage[T1]{fontenc}    
\usepackage{hyperref}       
\usepackage{url}            
\usepackage{booktabs}       
\usepackage{amsfonts}       
\usepackage{amsmath}
\usepackage{nicefrac}       
\usepackage{microtype}      

\begin{document}
	
	\title{Model adaptation and unsupervised learning with non-stationary batch data under smooth concept drift}
	
	\author{\name Subhro Das \email subhro.das@ibm.com \\
		\addr IBM T. J. Watson Research Center,
		Yorktown Heights, NY 10598, USA
		\AND
		\name Prasanth Lade \email Prasanth.Lade@us.bosch.com \\
		\addr Robert Bosch LLC,
		Palo Alto, CA 94304, USA
		\AND
		\name Soundar Srinivasan \email Soundar.Srinivasan@us.bosch.com \\
		\addr Robert Bosch LLC,
		Palo Alto, CA 94304, USA }
	

\maketitle

\begin{abstract} 
Most predictive models assume that training and test data are generated from a stationary process. However, this assumption does not hold true in practice. In this paper, we consider the scenario of a gradual concept drift due to the underlying non-stationarity of the data source. While previous work has investigated this scenario under a supervised-learning and adaption conditions, few have addressed the common, real-world scenario when labels are only available during training. We propose a novel, iterative algorithm for unsupervised adaptation of predictive models. We show that the performance  of our batch adapted prediction algorithm is better than that of its corresponding unadapted version. The proposed algorithm provides similar (or better, in most cases) performance within significantly less run time compared to other state of the art methods. We validate our claims though extensive numerical evaluations on both synthetic and real data.    
\end{abstract}

\section{Introduction}
\label{introduction}

It is crucial to adapt predictive models built on live non-stationary data streams to the ever changing underlying processes. Timely adaptation of these models to this phenomenon, known as concept drift, has drawn significant attention in contemporary literature. Adaptation to concept drift has a wide range of applications from Internet-of-Things (IoT) analytics to analysis of signals generated by autonomous robots, from spam detection to natural language processing. 

\begin{figure*}[t]
	\label{fig:drift}
	\vskip -0.2in
	\begin{center}
		\includegraphics[scale=0.41]{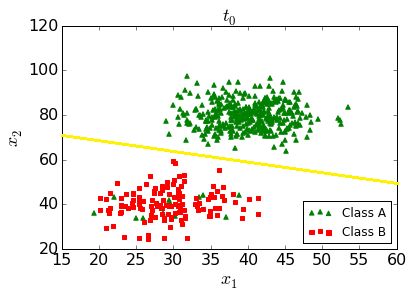} ~
		\includegraphics[scale=0.41]{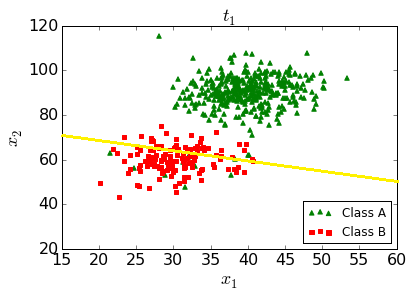} ~
		\includegraphics[scale=0.41]{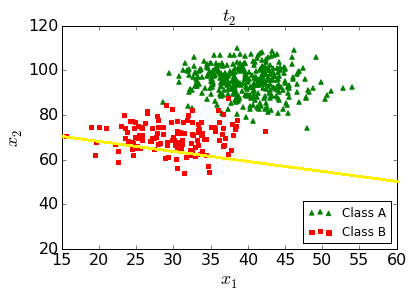}
		\caption{\footnotesize An example demonstrating the need of classifier adaptation when the data is undergoing gradual concept drift. }
		\vskip -0.1in
		\hrule
	\end{center}
	\vskip -0.3in
\end{figure*}

Most of the predictive models today operate under the assumption of a stationary environment. However certain real-time data, for example,  financial, climate, medical, energy demand, and pricing data, are generated from underlying non-stationary sources which are constantly changing with time. In Figure~\ref{fig:drift}, we demonstrate the effect of concept drift on prediction models with a visual illustration. In Figure~\ref{fig:drift}, we see that the prediction accuracy of the classifier (Classes A and B) degrades since the data (features $x_1$ and $x_2$) is undergoing smooth concept drift. So it is imperative to adapt the prediction models to new incoming data stream. 

In the manufacturing domain, there is a need to deploy predictive models to perform predictive maintenance, quality assessment and condition monitoring. But changes in either the machine configuration or their calibration or thresholds for quality assessment are the usual sources of concept drift in the data. Thus there is a need to adapt the deployed models to this gradual drift in the data. 

\begin{figure*}[t]
	\label{fig:methodology}
		\vskip -0.2in
	\begin{center}
		\includegraphics[scale=0.3]{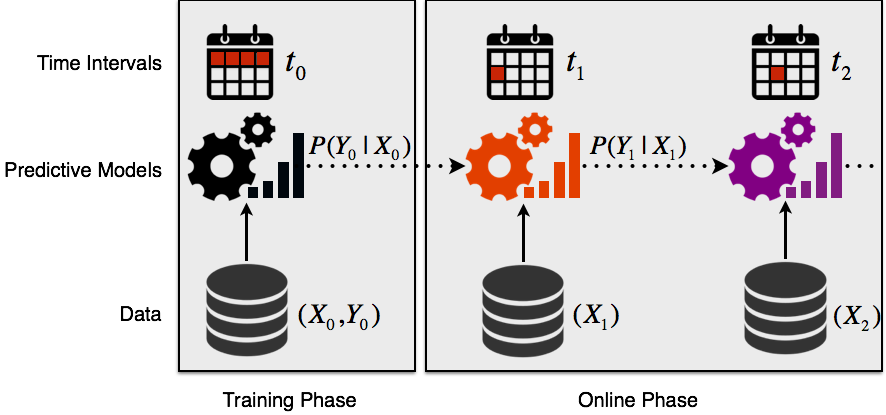}
		\caption{A brief description of the Model Adaptation Methodology.}
		\vskip 0.05in
		\hrule
	\end{center}
	\vskip -0.2in
\end{figure*}

The task of model adaptation becomes especially challenging when the incoming data streams are unlabeled or very sparsely labeled. Figure~\ref{fig:methodology} illustrates a specific use case where the features and labels, collected over a period of time, are available to learn a predictive model which is the Training phase ($t_0$). But once the model is deployed, we only have access to the trained model, $P(Y_0|X_0)$, given a training set $X_0$, but not the training data any more. At time $t_1$ when the model is online, it is pertinent to make prediction over a batch of data, $X_1$, by adapting the previously trained model $P(Y_0|X_0)$. We may not have access to the labels, $Y_1$, during adaptation and thus this presents a motivation to perform model adaptation with unsupervised learning on batch data. 

Although model adaptation has gained a lot of attention in the recent past, little has been done in developing a quantitative definition of concept drift. In this paper, we propose a novel quantitative expression of drifts of each of the data points in terms of changes in posterior probability distributions. We then develop a novel iterative algorithm that learns from the non-stationary data, estimates the point-wise drifts and adapts the prediction model to improve its accuracy. We evaluate the proposed algorithm on synthetic and real data, and, show significant improvement over an un-adapted solution.

The rest of the paper is organized as follows. In Section~\ref{sec:related}, we review some of the existing related literature and draw comparisons with our work. Section~\ref{sec:drift} formulates a quantitative definition of concept drift. The drift adaptation algorithm with its convergence properties are presented in Section~\ref{sec:model}. Section~\ref{sec:evaluation} includes the evaluation and results of the algorithm on synthetic and real data. We summarize and conclude in Section~\ref{sec:conclusion}.

\section{Related Work}
\label{sec:related}

Various approaches to address different problems in concept drift are summarized by \cite{moreno2012unifying}, \cite{gama2014survey}, \cite{heywood2015evolutionary}, \cite{ditzler2015learning} and \cite{vzliobaite2016overview}. \cite{moreno2012unifying} present a unifying framework to review and compare works in four different categories of dataset shifts. \cite{gama2014survey} cover various aspects of concept drift in an integrated way to reflect on the existing state of the art techniques and specifically focus on supervised learning. \cite{heywood2015evolutionary} survey developments in model building under both evolutionary and non-evolutionary streaming environments. \cite{vzliobaite2016overview} compile potential applications of concept drift adaptation to financial, climate, medical, energy demand and pricing data based on tasks, characteristics of changes and operational settings. \cite{kuznetsov2016time}, \cite{mohri2012new} present a series of models for time-series prediction under non-stationary environment.

In some applications, it is also crucial to detect when the data has undergone significant drift such that the existing model is no longer valid. Such goals leads to drift detection and it is relevant to batch adaptation. \cite{wang2015concept} present Linear Four Rates (LFR) framework to detect concept drift and identify the data points that belong to the new concept. However this detection technique is supervised and can be used only with binary classification models. \cite{dries2009adaptive} propose three methods for adaptive concept drift detection where the test statistics are dynamically adapted with the non-stationary data. Such methods turn out to be useful when the drift affects different characteristics of the underlying distribution at different time points.  

The primary objective of this paper is to continuously adapt the prediction model undergoing smooth/gradual concept drift, see \cite{bartlett1992learning}, irrespective of the degree of drift at a time point. \cite{dyer2014compose} introduce a computational geometry based framework to learn from non-stationary data, where labels are unavailable after initialization. They define non-stationarity in terms of time-varying probability distribution of the features, i.e., $P_t(x)$. They consider gradual drift, change in non-stationarity, as translation, rotation or compaction of $P_t(x)$. However, they do not quantify the drift. \cite{hanneke2015learning} study the bounds on the error rates of a predictive model given sequence of independent data points under concept drift. Further the paper provides an adaption method of active learning type where the bound on the number of labels, to achieve a desired bound on error rate, is studied. 
\cite{kuznetsov2016time} presents theoretical guarantees of ensemble based methods for forecasting non-stationary time-series. \cite{chaudhuri2012online} proposed a tracking algorithm where the observations follow a slightly drifted distribution. \cite{bousquet2002tracking} and \cite{herbster1998tracking} provide loss bounds of online algorithms for tracking the best set of experts, and extended them to shifting bounds in \cite{herbster2001tracking} for shifting predictors. \cite{kuznetsov2014generalization} and \cite{kuznetsov2015learning} presents time-series prediction error bounds for non-stationary mixing and non-mixing stochastic processes.

We are particularly interested in an unsupervised adaptation technique where the predictive model does not have access to the labels. In this context, \cite{hofer2013drift} presents an unsupervised statistical methodology for analyzing population drift in classification. They define non-stationarity in terms of time-varying probability distribution of the priors, i.e., $P_t(y)$ but the conditional feature distribution/density, $f(x | y)$, is stationary, where $x$ are the features and $y$ are the classes. They define drift as the ratio of $P_{t+1}(y)$ and $P_t(y)$. Such definitions of non-stationarity and drifts are restrictive, as in most practical applications the conditional feature distribution also undergoes gradual changes. We overcome this limitation and formulate an ubiquitous quantitative definition of concept drift, in the following Section \ref{sec:drift}, that encompasses more general non-stationarity in data distributions.

References~\cite{bach2010bayesian} and~\cite{kolter2007dynamic} have defined the concept drift in terms of the KL Divergence between the overall posterior probability distributions. In this paper, we quantify the point-wise drift, i.e., the change in the posterior probability of each data point, in terms of a physical quantity similar to the KL Divergence. Further, we estimate the point-wise drift in an unsupervised manner to improve the prediction accuracy. We point out that learning under non-stationary environment (concept drift) is different from domain adaptation~\cite{jiang2007instance}, \cite{cortes2014domain} and/or transfer learning~\cite{long2014adaptation}. Both in domain adaptation and transfer learning, the learning is done on source data (one distribution) and the prediction is done on a different/related target data (different distribution). Whereas, in this work we address a different problem setup, where the incoming data stream is being generated by a gradually drifting distribution as considered in \cite{souzaSDM:2015}, \cite{hofer2015adapting} and \cite{long2014adaptation}. There may or may not be any change in distribution between two subsequent batches of data.

%

\section{Drift Formulation}
\label{sec:drift}

Concept drift is the change in the statistical properties of the target variable over time and Fig.~\ref{fig:drift}--\ref{fig:methodology} visually portray the qualitative definition of concept drift. In this paper, we define drift in the context of classification problems. To obtain a quantitative definition of drift, we observe the changes in the joint and conditional probability distributions/density of the features and classes. 
%
%
\subsection{Problem Setup}
\label{subsec:problem}
We represent the data in terms of the predictor variables $\bf{X} \in \Omega$, where $\bf{\Omega}$ is the feature space, and the class labels $Y \in \{0,1\}$. The data can be represented by the joint probability distribution $f_{XY}({\bf{X}},Y)$. In most machine learning and data mining applications, predictive models are designed assuming that $f_{XY}({\bf{X}},Y)$ is time-invariant. However, in practice $f_{XY,t}({\bf{X}},Y)$ gradually changes with time, $t$, thereby causing concept drift. Since $f_{XY,t}({\bf{X}},Y)$ can be expressed as
\setlength\abovedisplayskip{4pt}
\setlength\belowdisplayskip{4pt}
\begin{small}
\begin{align}
\label{eqn:jointdist}
f_{XY,t}({\bf{X}},Y) = P_{Y|X,t}(Y=y | {\bf{X}} = {\bf{x}}) f_{X,t}({\bf{X}} = {\bf{x}})  = P_{Y,t}(Y=y) f_{X|Y,t}({\bf{X}} = {\bf{x}}| Y=y) ,
\end{align}
\end{small}
the drift can be sufficiently described by the time-varying nature of:
\begin{align}
P_{Y|X,t}(Y | {\bf{X}}) \;\; \text{AND} \;\; f_{X,t}({\bf{X}}) \qquad \qquad 
\text{OR}, \qquad\qquad  f_{X|Y,t}({\bf{X}} | Y) \;\; \text{AND} \;\; P_{Y,t}(Y). \nonumber
\end{align}
For notational simplicity, from now onwards we drop the subscripts $XY, X, Y, X|Y$ and $Y|X$ in the probability functions. In our formulation, we consider $P_t(Y | {\bf{X}})$ and $f_t({\bf{X}})$ for two reasons: $(i)$ for a given data point $\bf{x}$, its class label, $\widehat{y}$, is computed using the posterior probability $P_t(Y | {\bf{X}})$
\begin{align}
\label{eqn:classification}
\widehat{y} = \arg \max_{y} P_t(Y = y | {\bf{X} = x}),
\end{align}
and, $(ii)$ for unsupervised model adaptation at any time point we only observe the features, $\bf{x}$, which can provide us an estimate of feature distribution $f_t({\bf{X}})$.
\begin{remark}
	In this paper we study unsupervised model adaptation under concept drift. We aim to update the prediction rule based on the changes observed in the predictor variables~{\bf{X}} only. To be able to detect, and hence estimate, drift in the underlying model based on the variables~{\bf{X}} alone, we must observe some drift or changes in the feature distribution~$f_t({\bf{X}})$ with time. Otherwise, drift detection and adaptation would require labels~$Y$ of atleast some of the data points. Since in this paper we consider unsupervised learning, we assume that the feature distribution $f_t({\bf{X}})$ changes whenever there is a concept drift. 
\end{remark}
%
%

At time $t=0$, based on the labeled initial data (training phase in Figure~\ref{fig:methodology}) $({\bf{X}}^0, \bf{y}^0) = { ({\bf{x}}^0_1, y^0_1), \cdots ({\bf{x}}^0_{n_0}, y^0_{n_0}) }$, we learn the posterior distribution $P_0(y|{\bf{x}})$ and the feature density $f_0({\bf{x}})$. The underlying model $P_0(y|{\bf{x}})$ and $f_0({\bf{x}})$ are saved, but we do not store the initial training data. Then at time $t=1$ (online phase in Figure~\ref{fig:methodology}), we receive drifted unlabeled data ${\bf{x}}^1 = \{ {\bf{x}}^1_1, \cdots, {\bf{x}}^1_i, \cdots, {\bf{x}}^1_{n_1} \}$. Based on $P_0(y|{\bf{x}})$, $f_0({\bf{x}})$ and the new drifted data ${\bf{x}}^1$, we aim to compute the updated posterior distribution $P_1(y|{\bf{x}})$ and feature density $f_1({\bf{x}})$, keep them and discard the data ${\bf{x}}^1$. We repeat this process for every subsequent time step $t>=2$.  

Hence, at any time $t+1$, we have the old model $P_t(y|{\bf{x}})$ and $f_t({\bf{x}})$, and receive the new drifted data ${\bf{x}}^{t+1} = \{ {\bf{x}}^{t+1}_1, \cdots, {\bf{x}}^{t+1}_i, \cdots, {\bf{x}}^{t+1}_{n_{t+1}} \}$. Our objective is to design a method to obtain the updated $P_{t+1}(y|{\bf{x}})$ and $f_{t+1}({\bf{x}})$. This method will update the classifier at all time points. For simplicity of notation, we denote ${\bf{x}}^{t+1}_i$ by ${\bf{x}}_i$. To compute the updates, we first need to estimate the drifts which we define in the next subsection.

%
%
\subsection{Drift definition}
\label{subsec:definition}
We define point-wise drift, $\delta_t(y_i|{\bf{x}}_i)$, of each new data points ${\bf{x}}_i$ for each class $y_i$ at each time, 
\begin{align}
\label{eqn:drift}
\delta_t(y_i|{\bf{x}}_i) = P_t (y_i|{\bf{x}}_i) \log \frac{P_t (y_i|{\bf{x}}_i)}{P_{t+1}(y_i|{\bf{x}}_i)}.
\end{align}
From time $t$ to $t+1$, the change/divergence in posterior probability of data point ${\bf{x}}_i$ being in class~$y_i$ is captured in the drift, $\delta_t(y_i|{\bf{x}}_i)$. We use Kullback--Leibler (KL) divergence to measure the difference between two probability distributions. The conditional KL divergence of $P_{t+1}(y|{\bf{x}})$ from $P_t (y|{\bf{x}})$ is   
\begin{align}
\label{eqn:KL}
D \left( P_t \left(y|{\bf{x}}\right) || P_{t\!+\!1}\left(y|{\bf{x}}\right) \right) =  \sum_{{\bf{x}}_i \in {\bf{x}}} P_t ({\bf{x}}_i) \sum_{y_i \in y}  P_t (y_i|{\bf{x}}_i) \log \frac{P_t (y_i|{\bf{x}}_i)}{P_{t\!+\!1}(y_i|{\bf{x}}_i)}
\end{align}
where, $P_t ({\bf{x}}_i) = \frac{f_t ({\bf{x}}_i)}{\sum_{{\bf{x}}_i \in {\bf{x}}} f_t ({\bf{x}}_i)}$. The point-wise drifts defined in \eqref{eqn:drift} are the building blocks of the conditional KL divergence $D \left( P_t \left(y|{\bf{x}}\right) || P_{t\!+\!1}\left(y|{\bf{x}}\right) \right)$. Using \eqref{eqn:drift} and \eqref{eqn:KL}, we establish the relation
\begin{align}
\label{eqn:relationKL}
D \left( P_t \left(y|{\bf{x}}\right) || P_{t\!+\!1}\left(y|{\bf{x}}\right) \right) = \sum_{{\bf{x}}_i} \sum_{y_i} P_t ({\bf{x}}_i) \delta_t(y_i|{\bf{x}}_i). 
\end{align}
The KL divergence $D(: || :)$, is a non-negative quantity. A zero value denotes that there is no drift and higher the value stronger is the drift. We have developed a quantitative definition for drift which is in well accord with its qualitative notion. The overall drift between time points $t$ and $t+1$ can be expressed by the KL divergence of the discretized joint distributions $P_t({\bf{x}}, y)$ and $P_{t+1}({\bf{x}}, y)$,
\begin{align}
\label{eqn:jointKL}
D \left(P_t({\bf{x}}, y) || P_{t+1}({\bf{x}}, y) \right) = D \left(P_t({\bf{x}}) || P_{t+1}({\bf{x}}) \right) + D \left( P_t \left(y|{\bf{x}}\right) || P_{t\!+\!1}\left(y|{\bf{x}}\right) \right).
\end{align}
Any non-zero drift $D \left(P_t({\bf{x}}, y) || P_{t+1}({\bf{x}}, y)\right)$, is reflected in the divergence between the feature distributions, $D \left(P_t({\bf{x}}) || P_{t+1}({\bf{x}}) \right)$. The cases where the drift $D \left(P_t({\bf{x}}, y) || P_{t+1}({\bf{x}}, y) \right)$ results only in the changes of posterior probabilities, i.e., $D \left( P_t \left(y|{\bf{x}}\right) || P_{t\!+\!1}\left(y|{\bf{x}}\right) \right) > 0$, but no changes in the feature distribution, i.e., $D \left(P_t({\bf{x}}) || P_{t+1}({\bf{x}}) \right) = 0$, is beyond the scope of this paper. 
%
%
\subsection{Prediction using drift estimates}
\label{subsec:estimation}
We first estimate the point-wise drifts, $\delta_t(y_i|{\bf{x}}_i)$. The updated posterior probabilities, $P_{t+1}(y_i|{\bf{x}}_i)$, follow from the estimated drifts~$\delta_t(y_i|{\bf{x}}_i)$ using~\eqref{eqn:drift} as: 
\begin{align}
\label{eqn:posterior}
P_{t+1}(y_i|{\bf{x}}_i) = P_t (y_i|{\bf{x}}_i) e^{- \frac{ \delta_t(y_i|{\bf{x}}_i) }{P_t (y_i|{\bf{x}}_i)} }.
\end{align}
The drift $\delta_t(y_i|{\bf{x}}_i)$ captures the divergence of the posterior probability of the data point~${\bf{x}}_i$ belonging to class~$y_i$ from time~$t$ to time~$t+1$. Once we obtain the estimated posterior probabilities~$\widehat{P}_{t+1}(y_i|{\bf{x}}_i)$ using the drift estimates, we predict the class of data point ${\bf{x}}_i$ as:
\begin{align}
\label{eqn:class}
\widehat{y}_i = \arg \max_{y_i} \widehat{P}_{t+1}(y_i|{\bf{x}}_i).
\end{align}
We see that the class labels~$y_i$ are the maximum a posteriori probability (MAP) estimates, which, in turn, are dependent on the drift estimates~$\delta_t(y_i|{\bf{x}}_i)$. In the next section, we propose an iterative algorithm that simultaneously estimates the drifts and the posterior probabilities of each data point~${\bf{x}}_i$.

\section{Model Adaptation}
\label{sec:model}
\begin{algorithm}[t]
	\caption{Model adaptation with estimated drifts}
	\label{alg:algorithm}
	\begin{algorithmic}
		\STATE {\bfseries Input:} model $P_t(y|x)$, data $\bf{X} = \{x_i\}$   
		\vskip2pt
		\STATE {\bfseries Initialize:} $\widehat{P}^{(0)}_{t+1}(y|x_i) = P_t(y|x_i)$, $\widehat{\delta}^{(0)}(y|x_i) = 0$.
		\vskip2pt
		\FOR{$k=0$ {\bfseries to} MaxIterations$-1$}
		\vskip2pt
		\STATE $\widehat{y}^{(k)}_{i} = \displaystyle \max_{y} \widehat{P}^{k}_{t+1}(y|x_i)$
		\vskip2pt
		\STATE Using $\{(x_i, \widehat{y}^{(k)}_{i})\}$, compute $\overline{P}^{(k)}_{t+1}(y|x_i)$
		\vskip2pt
		\STATE $\textit{KL}^{(k)} = D( \widehat{P}^{(k)}_{t+1}(y|x_i) || \overline{P}^{(k)}_{t+1}(y|x_i) )$
		\vskip2pt
		\IF{$\textit{KL}^{(k)} >$ tolerance} 
		\vskip2pt
		\STATE $\widehat{\delta}^{(k+1)}(y|x_i) = \widehat{\delta}^{k}(y|x_i) + \gamma^{(k)}_i \Delta^{(k)}_i$, $\;$ where, $\Delta^{(k)}_i =$ - grad $\{\textit{KL}^{(k)}\}$, $\gamma^{(k)}_i =$ step size  
		\vskip2pt
		\STATE $\widehat{P}^{(k+1)}_{t+1}(y|x_i) = P_t(y|x_i) e^{- \frac{ \widehat{\delta}^{(k+1)}(y|x_i) }{P_t(y|x_i)} }$
		\vskip2pt
		\STATE $\!\!\!\!\!\!\!$ {\bfseries else} exit for loop
		\ENDIF
		\ENDFOR 
		\vskip2pt
		\STATE {\bfseries Return:} $\widehat{P}_{t+1}(y_i|{\bf{x}}_i) = \widehat{P}^{(k)}_{t+1}(y|x_i)$.
	\end{algorithmic}
\end{algorithm}
In this section, we propose the novel model adaptation algorithm for unsupervised learning of class labels from non-stationary data under concept drift.  

\subsection{Algorithm}
The unsupervised algorithm for model adaptation runs two companion posterior probability estimation sub-routines at each iteration,~$k = 0, \cdots, $ until it converges. In the first subroutine we update the point-wise drift estimates~$\widehat{\delta}^{k}(y|x_i)$ and then update the drift-based posterior probability estimates~$\widehat{P}^{(k)}_{t+1}(y|x_i)$ using~\eqref{eqn:posterior}. The second sub-routine computes the class labels~$\widehat{y}^{(k)}_i$ from~\eqref{eqn:classification}, and then, using~$\{{\bf{x}}_i, \widehat{y}^{(k)}_i\}$, updates the label-based posterior probability estimates~$\overline{P}^{(k)}_{t+1}(y|x_i)$. The algorithm, proposed in this paper, iteratively decreases the divergence between the drift-based and label-based posterior probability estimates. The decreasing KL divergence is guaranteed by a gradient descent (derivative w.r.t drift) step on the drift updates~$\widehat{\delta}^{k}(y|x_i)$. The steps of the model adaptation iterations are presented in Algorithm~\ref{alg:algorithm}.

In the algorithm, the drift-based posterior $\widehat{P}^{(0)}_{t+1}(y|x_i)$ is initialized with the posterior distribution, $P_t(y|x_i)$, from previous time step $t$ and the drift estimates are initialized to be~$0$. At each iteration, the labels $\widehat{y}^{(k)}_{i}$ and thereafter $\overline{P}^{(k)}_{t+1}(y|x_i)$ are estimated using $\widehat{P}^{(k)}_{t+1}(y|x_i)$ distribution obtained from the previous iteration. We compute the KL divergence between the label-based and drift-based posterior distributions, and test for convergence. If the divergence is greater than the pre-defined threshold, the gradient of the KL divergence is used to update the drifts and the corresponding posterior distribution. The algorithm continues until it reaches convergence. In the next Subsection~\ref{subsec:conv} we discuss the convergence properties and the design of the parameters of the Algorithm~\ref{alg:algorithm}. 
 
\subsection{Convergence Properties and Step Size Design}
\label{subsec:conv}
In model adaptation Algorithm~\ref{alg:algorithm}, the cost function is the KL divergence~$\textit{KL}^{(k)}$ between the two posterior probabilities,~$\widehat{P}^{(k)}_{t+1}(y|x_i)$ and~$\overline{P}^{(k)}_{t+1}(y|x_i)$. Since, KL divergence is a convex function and we are minimizing it using a gradient descent step, the algorithm monotonically converges due to the convex property of the cost function. Now the rate of convergence and the asymptotic bounds depend on the choice of the step size. Hence, the key design parameter in model adaptation Algorithm~\ref{alg:algorithm} is the step size~$\gamma^{(k)}_i$. Here, we consider step size~$\gamma^{(k)}_i$ monotonically decreasing with iterations~$k$, i.e., 
\begin{align}
\label{eqn:stepsize}
\gamma^{(k)}_i = \frac{c}{\sqrt{k+1}}, \qquad \forall i, \;\; k = 0, 1, \cdots
\end{align}
where, $c$ is a constant. The dynamic and decreasing step size ensures that the algorithm achieves faster convergence rate and at the same time converges with a low KL divergence between the between the drift-based and label-based posterior probability estimates. Thus, the algorithm ensures that they converge to the true posterior probabilities of each data point. We evaluate and discuss the predictive performance of the proposed algorithm in the following section.     
 
\section{Evaluation and Results}
\label{sec:evaluation}
\begin{figure}[t]
	\begin{center}
		\includegraphics[scale=0.2]{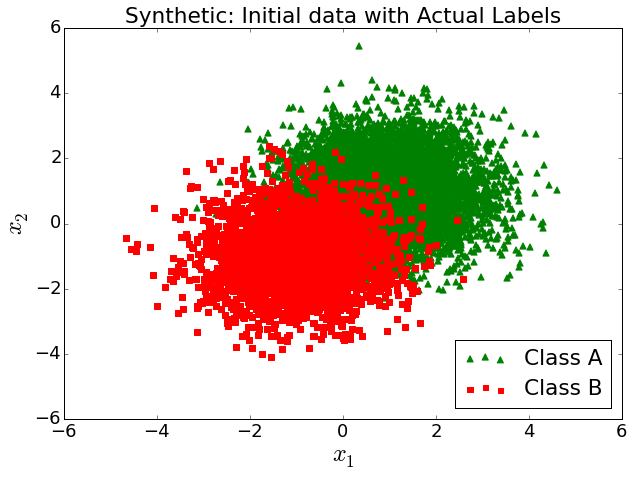} 
		\includegraphics[scale=0.2]{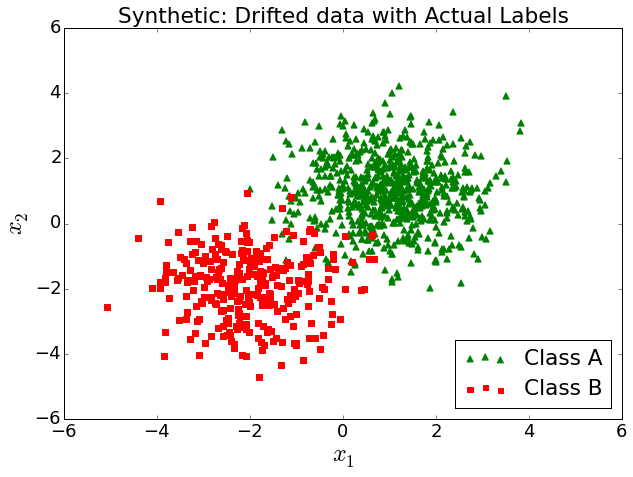} 
		\includegraphics[scale=0.2]{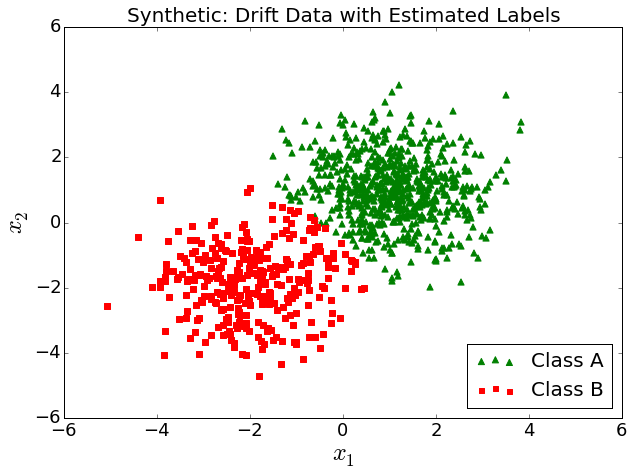}
		\includegraphics[scale=0.2]{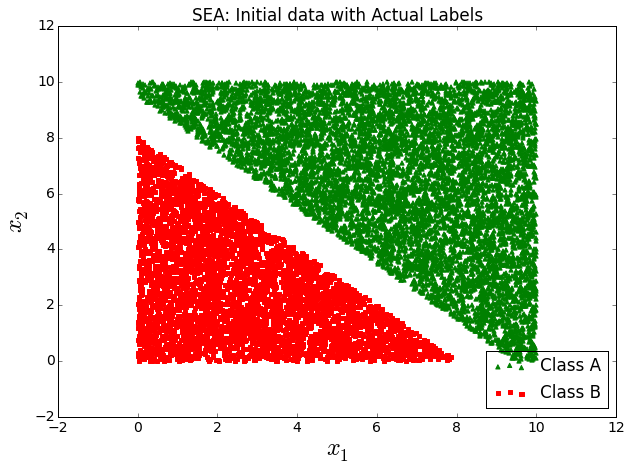} 
		\includegraphics[scale=0.2]{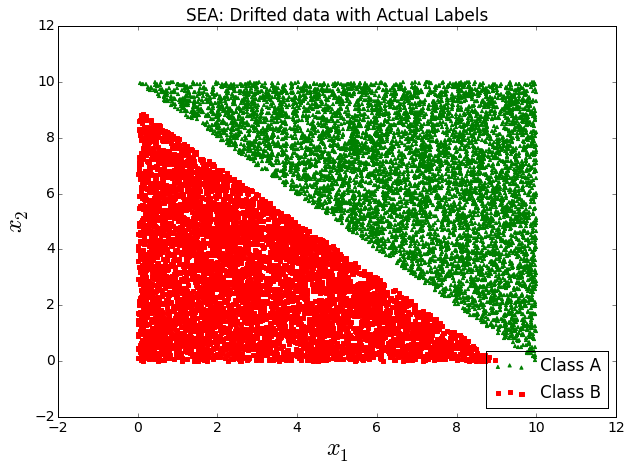}
		\includegraphics[scale=0.2]{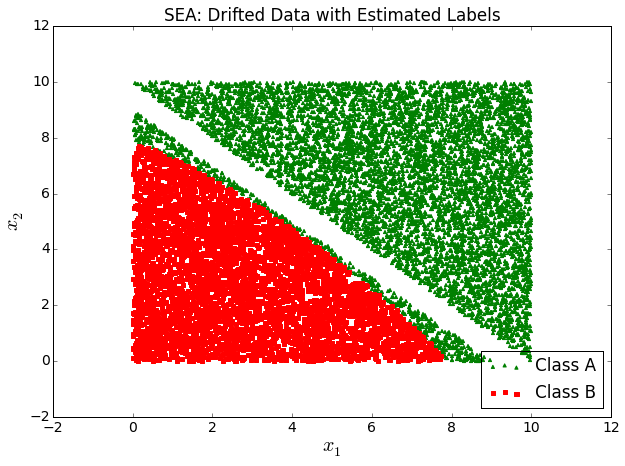}
		\caption{Initial and Drifted data with actual labels, and Drifted data with labels estimated using Algorithm~\ref{alg:algorithm} for Synthetic Data (Top) and Modified SEA Data (Bottom)}
	\end{center}
	\label{fig:synthetic_sea}
	\vskip-10pt
	\hrule
\end{figure}
We extensively evaluate the performance of the proposed algorithm on synthetic, SEA~(\cite{street2001streaming}), and manufacturing datasets. We benchmark our solution with respect to supervised and unadapted approaches. In the supervised approach, we train and test a new model using the features and labels of the drifted data at time $t+1$ by $10$-fold cross validation. Although we have the constraint that the labels are not available, we evaluate our model against supervised model to assess the benchmark performance. The unadapted model uses the model learnt at time $t$ to predict the labels at time $t+1$. The performance of a supervised learning solution is the best case scenario, whereas the unadapted learning is the worst case scenario. We empirically demonstrate that the performance of Algorithm~\ref{alg:algorithm} improves on the unadapted solution, but yields a performance gap compared to the supervised learning. To further reduce this gap is the scope for future work.
%
\subsection{Evaluation on synthetic dataset}
\label{subsec:synthetic} 
The synthetic data is a mixture of two Gaussians, i.e., ${\bf{x}}^t | y^t \sim \mathcal{N} \left( \mu^t_y, \Sigma^t_y \right)$, where ${\bf{x}}^t = \left[x^t_1 \;\; x^t_2\right]^T$ and $y = \{0,1\}$. The prior on the class labels is considered to follow Bernoulli distribution, i.e., $y^t \sim $ Bernoulli$(p_t)$. The prior probability, mean and covariances of the Gaussian distributions for the initial $(t)$ and drifted $(t+1)$ data are:
\begin{small}
\begin{align}
p_t = p_{t+1} = 0.7; \mu^t_0 = \mu^{t+1}_0 = \begin{bmatrix} 1 \\ 1 \end{bmatrix}; \mu^t_1 = \begin{bmatrix} -1 \\ -1 \end{bmatrix}; \mu^{t+1}_1 = \begin{bmatrix} -2 \\ -2 \end{bmatrix}; \Sigma^t_y = \Sigma^{t+1}_y = \begin{bmatrix} 1 & 0 \\ 0 & 1 \end{bmatrix}\;\;\forall y = \{0,1\} \nonumber
\end{align}
\end{small}
We considered $N_t = 10,000$ initial data points and $N_{t+1} = 1,000$ drifted data points for our experimental evaluations. The drift in the mean of class $y=1$ data points from $\mu^t_1$ to $\mu^{t+1}_1$ is graphically displayed in Fig~3 (Top). We report the classification error for the drifted data in Table~\ref{table:performance} and observe that the prediction accuracy of the adapted model is higher than the unadapted model. The last plot in Fig~3 (Top) shows the labels of the drifted data estimated using Algorithm~\ref{alg:algorithm}.
\begin{table}[h]
	\vskip-10pt
	\caption{Performance on synthetic, SEA and manufacturing datasets} 
	\centering 
	\begin{tabular}{ | c | c | c | c | } 
		\hline  
		\rule{0pt}{3ex}  Classification errors & Synthetic & SEA & Manufacturing \\ [1ex] 
		\hline 
		\rule{0pt}{3ex} Supervised learning (best-case) & 2.20\% & 4.45\% & 1.40\% \\ [1ex]
		Model adaptation (Algo.~\ref{alg:algorithm}) & 3.10\% & 4.97\% & 1.67\% \\ [1ex]
		Without adaptation (worst-case) & 4.10\% & 5.61\% & 1.85\% \\ [1ex] 
		\hline 
	\end{tabular} 
	\label{table:performance}
	\vskip-10pt
\end{table}
%
\subsection{Evaluation on SEA dataset}
\label{subsec:SEA}
We adapted the SEA concepts dataset from~\cite{street2001streaming} and removed data points corresponding to one of the three classes. We changed the classification rule so as to incorporate the change in the feature distribution from the initial data to the drifted data. The initial and drifted data for the modified SEA dataset can be seen in Fig.~3 (Bottom). Similar to the results of the synthetic data, we see an improvement in the accuracy of the adapted model over the unadapted one, see Table~\ref{table:performance}. We have also plotted the estimated labels for the drifted data in Fig.~3 (Bottom). We observe from this plot that the adapted model was able to estimate a class boundary that is much closer to the class boundary in the drifted data rather than the initial data.

\subsection{Evaluation on Manufacturing dataset}
\label{subsec:Manufacturing}
In this study we have evaluated our algorithm on a real dataset from a manufacturing plant. This data contains 27 different measurements taken from a product that is  manufactured in the plant and the quality of the end product (Good/Bad) is used as a class label. This is a binary classification problem and there is a gradual drift in the data feature space. We evaluated a total of 11,000 products that have been produced and divided them into two phases, training (initial data) and testing (drifted data). We applied our algorithm to this data and the results are chronicled in Table~\ref{table:performance}. Again in this case, we observe that the adapted model outperforms the unadapted model which is surpassed by the supervised model. In all of our above experiments, we have chosen step size empirically to be $\gamma^{(k)} = \frac{50}{\sqrt{k+1}}$, the convergence threshold to be~$10^{-10}$ and number of iterations as~$10$.


\subsection{Comparative Evaluation}
\label{subsec:Comparison}
\begin{figure}[t]
	\label{fig:comparison}
	\begin{center}
		\includegraphics[scale=0.5]{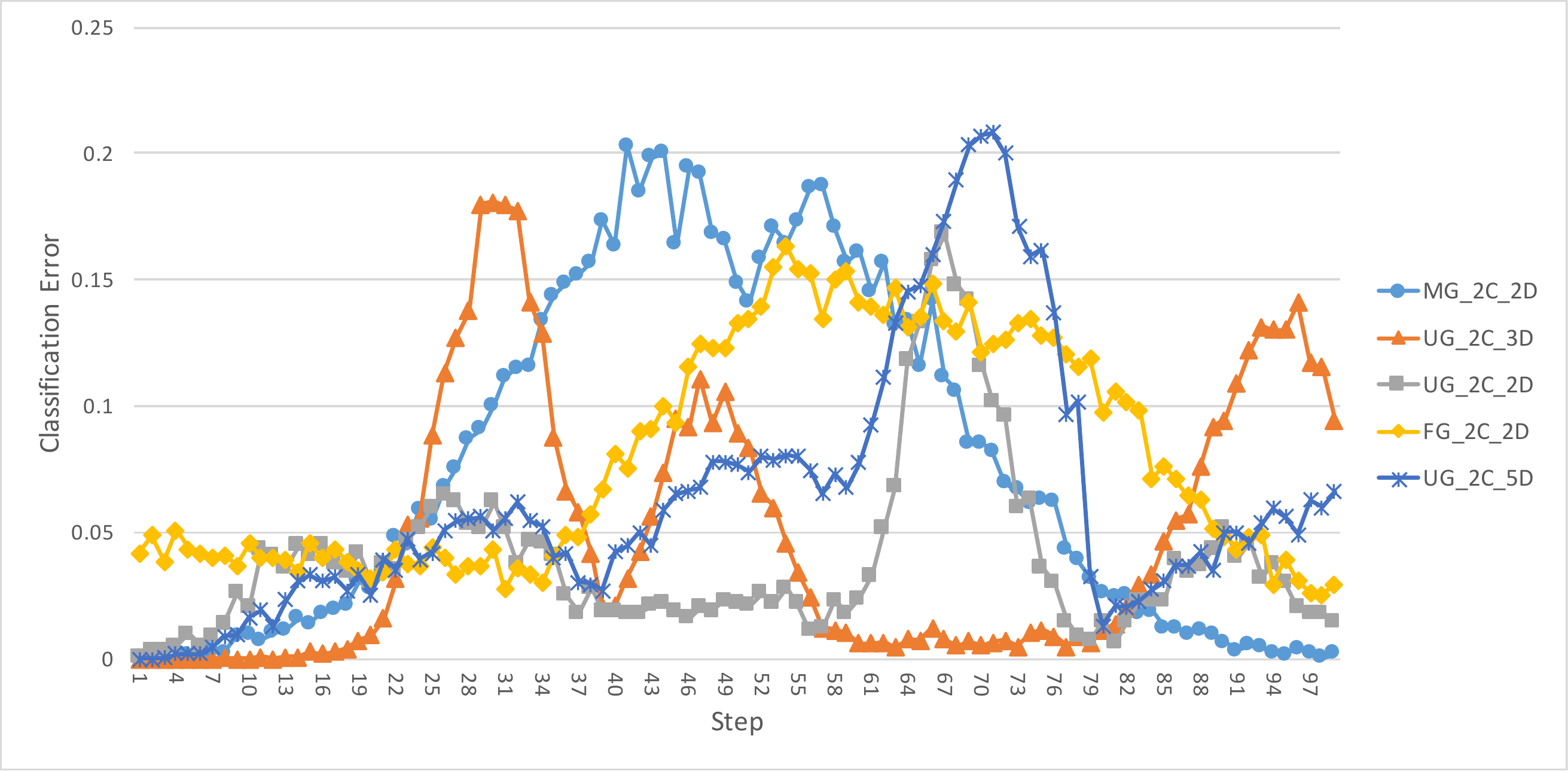}
		\caption{Classification error of Algorithm~\ref{alg:algorithm} with time on datasets from \cite{souzaSDM:2015}.}
		\hrule
	\end{center}
	\vskip -0.2in
\end{figure}
Here, we compare the performance of our model adaptation algorithm~\ref{alg:algorithm} with the state of the art method {\it Stream Classification Algorithm Guided by Clustering} (SGARC) presented in~\cite{souzaSDM:2015}.  In Fig.~4, we present the classification error on the drifted data with time on the five datasets MG\_2C\_2D, UG\_2C\_3D, UG\_2C\_2D, FG\_2C\_2D and UG\_2C\_5D from~\cite{souzaSDM:2015}, out of which the first three datasets were originally proposed in~\cite{dyer2014compose}. \cite{souzaSDM:2015} compared the performance of their algorithm SGARC with {\it Compacted Object Sample Extraction} (COMPOSE) from~\cite{dyer2014compose} and {\it Arbitrary Sub-population Tracker} (APT) from~\cite{krempl2011algorithm}. \cite{souzaSDM:2015} demonstrated that SGARC outperforms both COMPOSE and APT on MG\_2C\_2D dataset, whereas on the UG\_2C\_2D dataset, SGARC outperforms APT while providing same performance as COMPOSE. 
\begin{table}[h]
	\vskip-10pt
	\caption{Performance comparison with SGARC from \cite{souzaSDM:2015}.} 
	\centering 
	\begin{tabular}{ | c | c | c | c | c | c | } 
		\hline  
		\rule{0pt}{3ex}  Classification accuracy & MG\_2C\_2D & UG\_2C\_3D & UG\_2C\_2D & FG\_2C\_2D & UG\_2C\_5D \\ [1ex] 
		\hline 
		\rule{0pt}{3ex} SGARC (1NN) & 92.71\% &	94.77\% &	95.56\% &	95.16\% &	90.98\%  \\ [1ex]
		SGARC (SVM) & 92.75\% &	94.79\% &	95.53\% &	95.23\% &	88.24\%  \\ [1ex]
		Our Model (Algo.~\ref{alg:algorithm}) & 92.20\% &	\bf{95.11\%} &	\bf{96.08\%} &	92.03\% &	\bf{93.65\%} \\ [1ex] 
		\hline 
	\end{tabular} 
	\vskip-10pt
	\label{table:sgarc}
\end{table}
In Table~\ref{table:sgarc}, we see that our method outperforms SGARC on UG\_2C\_2D, UG\_2C\_3D and UG\_2C\_5D datasets, whereas lags behind (but still comparative) on the MG\_2C\_2D and FG\_2C\_2D datasets. Note that, our method is a less computationally complex and converges within 10~iterations. Table~\ref{table:sgarc_time} exhibits that our method achieves comparative performance (see Table~\ref{table:sgarc}) within a significantly less time compared to SGARC. \cite{souzaSDM:2015} showed that SGARC runs much faster than the COMPOSE and APT on the same five datasets. Hence, our model adaptation algorithm~\ref{alg:algorithm} provides comparative (or better, in most cases) classification accuracy with significantly less computation run times. 
\begin{table}[h]
	\vskip-10pt
	\caption{Computation time comparison with SGARC from \cite{souzaSDM:2015}.} 
	\centering 
	\begin{tabular}{ | c | c | c | c | c | c | } 
		\hline  
		\rule{0pt}{3ex}  Time (in seconds) & MG\_2C\_2D & UG\_2C\_3D & UG\_2C\_2D & FG\_2C\_2D & UG\_2C\_5D \\ [1ex] 
		\hline 
		\rule{0pt}{3ex} SGARC (1NN) & 117.67 &	118.41 &	59.76 &	119.7 &	120.21  \\ [1ex]
		SGARC (SVM) & 21.37 &	20.78 &	10.11 &	20.25 &	31.18   \\ [1ex]
		Our Model (Algo.~\ref{alg:algorithm}) & \bf{5.20} &	\bf{2.50} &	\bf{5.27} &	\bf{5.20} &	\bf{5.50}  \\ [1ex] 
		\hline 
	\end{tabular} 
	\vskip-10pt
	\label{table:sgarc_time}
\end{table}
%

\section{Conclusions} 
\label{sec:conclusion}
Making classifiers robust to drift is an important requirement for practical applications. In this work, we have presented an example of model adaptation for scenarios where drift is gradual and labels are unavailable during the adaptation period. The three primary contributions of this paper are: (i) quantification of concept drift in classification applications; (ii) determination of sample importance in the estimation of drift; and (iii) development of a novel algorithm that estimates the drift of each data point and adapts the classifier to improve prediction accuracy. While the adaptation algorithm has shown promising results for a Naive-Bayes classifier, extensions to other, popular classifiers will have to be examined. Similarly, an extension from a current batch-implementation to a streaming-implementation is desired. Future work will also investigate the convergence of the algorithm in situations where the drift is not gradual. 
\vskip-20pt


\begin{small}

\bibliography{references}

\begin{thebibliography}{28}
\providecommand{\natexlab}[1]{#1}
\providecommand{\url}[1]{\texttt{#1}}
\expandafter\ifx\csname urlstyle\endcsname\relax
  \providecommand{\doi}[1]{doi: #1}\else
  \providecommand{\doi}{doi: \begingroup \urlstyle{rm}\Url}\fi

\bibitem[Bach and Maloof(2010)]{bach2010bayesian}
Stephen Bach and Mark Maloof.
\newblock A bayesian approach to concept drift.
\newblock In \emph{Advances in Neural Information Processing Systems}, pages
  127--135, 2010.

\bibitem[Bartlett(1992)]{bartlett1992learning}
Peter~L Bartlett.
\newblock Learning with a slowly changing distribution.
\newblock In \emph{Proceedings of the fifth annual workshop on Computational
  learning theory}, pages 243--252. ACM, 1992.

\bibitem[Bousquet and Warmuth(2002)]{bousquet2002tracking}
Olivier Bousquet and Manfred~K Warmuth.
\newblock Tracking a small set of experts by mixing past posteriors.
\newblock \emph{Journal of Machine Learning Research}, 3\penalty0
  (Nov):\penalty0 363--396, 2002.

\bibitem[Chaudhuri et~al.(2010)Chaudhuri, Freund, and Hsu]{chaudhuri2012online}
Kamalika Chaudhuri, Yoav Freund, and Daniel Hsu.
\newblock An online learning-based framework for tracking.
\newblock In \emph{Uncertainty in Artificial Intelligence (UAI)}, 2010.

\bibitem[Cortes and Mohri(2014)]{cortes2014domain}
Corinna Cortes and Mehryar Mohri.
\newblock Domain adaptation and sample bias correction theory and algorithm for
  regression.
\newblock \emph{Theoretical Computer Science}, 519:\penalty0 103--126, 2014.

\bibitem[Ditzler et~al.(2015)Ditzler, Roveri, Alippi, and
  Polikar]{ditzler2015learning}
Gregory Ditzler, Manuel Roveri, Cesare Alippi, and Robi Polikar.
\newblock Learning in nonstationary environments: A survey.
\newblock \emph{Computational Intelligence Magazine, IEEE}, 10\penalty0
  (4):\penalty0 12--25, 2015.

\bibitem[Dries and R{\"u}ckert(2009)]{dries2009adaptive}
Anton Dries and Ulrich R{\"u}ckert.
\newblock Adaptive concept drift detection.
\newblock \emph{Statistical Analysis and Data Mining}, 2:\penalty0 311--327,
  2009.

\bibitem[Dyer et~al.(2014)Dyer, Capo, and Polikar]{dyer2014compose}
Karl~B Dyer, Robert Capo, and Robi Polikar.
\newblock Compose: A semisupervised learning framework for initially labeled
  nonstationary streaming data.
\newblock \emph{IEEE Transactions on Neural Networks and Learning Systems},
  2014.

\bibitem[Gama et~al.(2014)Gama, {\v{Z}}liobait{\.e}, Bifet, Pechenizkiy, and
  Bouchachia]{gama2014survey}
Jo{\~a}o Gama, Indr{\.e} {\v{Z}}liobait{\.e}, Albert Bifet, Mykola Pechenizkiy,
  and Abdelhamid Bouchachia.
\newblock A survey on concept drift adaptation.
\newblock \emph{ACM Computing Surveys (CSUR)}, 46\penalty0 (4):\penalty0 44,
  2014.

\bibitem[Hanneke et~al.(2015)Hanneke, Kanade, and Yang]{hanneke2015learning}
Steve Hanneke, Varun Kanade, and Liu Yang.
\newblock Learning with a drifting target concept.
\newblock In \emph{Algorithmic Learning Theory}, pages 149--164. Springer,
  2015.

\bibitem[Herbster and Warmuth(1998)]{herbster1998tracking}
Mark Herbster and Manfred~K Warmuth.
\newblock Tracking the best expert.
\newblock \emph{Machine Learning}, 32\penalty0 (2):\penalty0 151--178, 1998.

\bibitem[Herbster and Warmuth(2001)]{herbster2001tracking}
Mark Herbster and Manfred~K Warmuth.
\newblock Tracking the best linear predictor.
\newblock \emph{Journal of Machine Learning Research}, 1\penalty0
  (Sep):\penalty0 281--309, 2001.

\bibitem[Heywood(2015)]{heywood2015evolutionary}
Malcolm~I Heywood.
\newblock Evolutionary model building under streaming data for classification
  tasks: opportunities and challenges.
\newblock \emph{Genetic Programming and Evolvable Machines}, 16\penalty0
  (3):\penalty0 283--326, 2015.

\bibitem[Hofer(2015)]{hofer2015adapting}
Vera Hofer.
\newblock Adapting a classification rule to local and global shift when only
  unlabelled data are available.
\newblock \emph{European Journal of Operational Research}, 243\penalty0
  (1):\penalty0 177--189, 2015.

\bibitem[Hofer and Krempl(2013)]{hofer2013drift}
Vera Hofer and Georg Krempl.
\newblock Drift mining in data: A framework for addressing drift in
  classification.
\newblock \emph{Computational Statistics \& Data Analysis}, 57\penalty0
  (1):\penalty0 377--391, 2013.

\bibitem[Jiang and Zhai(2007)]{jiang2007instance}
Jing Jiang and ChengXiang Zhai.
\newblock Instance weighting for domain adaptation in {NLP}.
\newblock \emph{ACL}, 7:\penalty0 264--271, 2007.

\bibitem[Kolter and Maloof(2007)]{kolter2007dynamic}
J~Zico Kolter and Marcus~A Maloof.
\newblock Dynamic weighted majority: An ensemble method for drifting concepts.
\newblock \emph{Journal of Machine Learning Research}, 8\penalty0
  (Dec):\penalty0 2755--2790, 2007.

\bibitem[Krempl(2011)]{krempl2011algorithm}
Georg Krempl.
\newblock The algorithm apt to classify in concurrence of latency and drift.
\newblock In \emph{International Symposium on Intelligent Data Analysis}, pages
  222--233. Springer, 2011.

\bibitem[Kuznetsov and Mohri(2014)]{kuznetsov2014generalization}
Vitaly Kuznetsov and Mehryar Mohri.
\newblock Generalization bounds for time series prediction with non-stationary
  processes.
\newblock In \emph{International Conference on Algorithmic Learning Theory},
  pages 260--274. Springer, 2014.

\bibitem[Kuznetsov and Mohri(2015)]{kuznetsov2015learning}
Vitaly Kuznetsov and Mehryar Mohri.
\newblock Learning theory and algorithms for forecasting non-stationary time
  series.
\newblock In \emph{Advances in Neural Information Processing Systems}, pages
  541--549, 2015.

\bibitem[Kuznetsov and Mohri(2016)]{kuznetsov2016time}
Vitaly Kuznetsov and Mehryar Mohri.
\newblock Time series prediction and online learning.
\newblock In \emph{29th Annual Conference on Learning Theory}, pages
  1190--1213, 2016.

\bibitem[Long et~al.(2014)Long, Wang, Ding, Pan, and
  Philip]{long2014adaptation}
Mingsheng Long, Jianmin Wang, Guiguang Ding, Sinno~Jialin Pan, and S~Yu Philip.
\newblock Adaptation regularization: A general framework for transfer learning.
\newblock \emph{IEEE Transactions on Knowledge and Data Engineering}, 2014.

\bibitem[Mohri and Medina(2012)]{mohri2012new}
Mehryar Mohri and Andres~Munoz Medina.
\newblock New analysis and algorithm for learning with drifting distributions.
\newblock In \emph{International Conference on Algorithmic Learning Theory},
  pages 124--138. Springer, 2012.

\bibitem[Moreno-Torres et~al.(2012)Moreno-Torres, Raeder, Alaiz-Rodr{\'\i}Guez,
  Chawla, and Herrera]{moreno2012unifying}
Jose~G Moreno-Torres, Troy Raeder, Roc{\'\i}O Alaiz-Rodr{\'\i}Guez, Nitesh~V
  Chawla, and Francisco Herrera.
\newblock A unifying view on dataset shift in classification.
\newblock \emph{Pattern Recognition}, 45\penalty0 (1):\penalty0 521--530, 2012.

\bibitem[Souza et~al.(2015)Souza, Silva, Gama, and Batista]{souzaSDM:2015}
V.~M.~A. Souza, D.~F. Silva, J.~Gama, and G.~E. A. P.~A. Batista.
\newblock Data stream classification guided by clustering on nonstationary
  environments and extreme verification latency.
\newblock In \emph{Proceedings of SIAM International Conference on Data Mining
  (SDM)}, pages 873--881, 2015.

\bibitem[Street and Kim(2001)]{street2001streaming}
W~Nick Street and YongSeog Kim.
\newblock A streaming ensemble algorithm (sea) for large-scale classification.
\newblock In \emph{International conference on knowledge discovery and data
  mining (SIGKDD)}, pages 377--382. ACM, 2001.

\bibitem[Wang and Abraham(2015)]{wang2015concept}
Heng Wang and Z.~Abraham.
\newblock Concept drift detection for streaming data.
\newblock In \emph{2015 International Joint Conference on Neural Networks
  (IJCNN)}, pages 1--9, July 2015.

\bibitem[{\v{Z}}liobait{\.e} et~al.(2016){\v{Z}}liobait{\.e}, Pechenizkiy, and
  Gama]{vzliobaite2016overview}
Indr{\.e} {\v{Z}}liobait{\.e}, Mykola Pechenizkiy, and Jo{\~a}o Gama.
\newblock An overview of concept drift applications.
\newblock In \emph{Big Data Analysis: New Algorithms for a New Society}, pages
  91--114. Springer, 2016.

\end{thebibliography}
\end{small}

\end{document}